\documentclass[10pt,twocolumn,letterpaper]{article}

\usepackage{cvpr}
\usepackage{times}
\usepackage{epsfig}
\usepackage{graphicx}
\usepackage{amsmath}
\usepackage{amssymb}

\usepackage{multirow}

\usepackage{amsbsy}
\usepackage{mathtools}

\usepackage{bm}
\bmdefine\bmu{\mu}
\bmdefine\bSigma{\Sigma}
\bmdefine\btheta{\Theta}
		
\usepackage{mathtools}

\usepackage{accents}

\DeclareMathOperator*{\argmax}{argmax}

\DeclarePairedDelimiter{\ceil}{\lceil}{\rceil}

\usepackage[breaklinks=true,bookmarks=false]{hyperref}

\cvprfinalcopy 

\def\cvprPaperID{****} 

\begin{document}

\title{Point-to-Pose Voting based Hand Pose Estimation using Residual Permutation Equivariant Layer}

\author{Shile Li\\
Technical University of Munich\\
{\tt\small li.shile@mytum.de}
\and
Dongheui Lee\\
Technical University of Munich,\\
German Aerospace Center\\
{\tt\small dhlee@tum.de}
}

\maketitle

\begin{abstract}
Recently, 3D input data based hand pose estimation methods have shown state-of-the-art performance, because 3D data capture more spatial information than the depth image.
Whereas 3D voxel-based methods need a large amount of memory, PointNet based methods need tedious preprocessing steps such as K-nearest neighbour search for each point.
In this paper, we present a novel deep learning hand pose estimation method for an unordered point cloud.
Our method takes 1024 3D points as input and does not require additional information.
We use Permutation Equivariant Layer (PEL) as the basic element, where a residual network version of PEL is proposed for the hand pose estimation task.
Furthermore, we propose a voting-based scheme to merge information from individual points to the final pose output. 
In addition to the pose estimation task, the voting-based scheme can also provide point cloud segmentation result without ground-truth for segmentation.
We evaluate our method on both NYU dataset and the Hands2017Challenge dataset. Our method outperforms recent state-of-the-art methods, where our pose accuracy is currently the best for the Hands2017Challenge dataset.

\end{abstract}

\vspace{-0.4cm}
\section{Introduction}
\label{sec:introduction}

%

Hand pose estimation plays an important role in human-robot interaction tasks, such as gesture recognition and learning grasping capability by human demonstration. 
Since the emergence of consumer level depth sensing devices, a lot of depth image based hand pose estimation methods appeared.
Many state-of-the-art methods use depth image as input, which provides the conveniences to use the well developed convolutional neural networks or residual networks.
However, methods using 2D images as input cannot fully utilize 3D spatial information in the depth image.
Furthermore, the appearance of the depth image is dependent on the camera parameters, such that the trained model using one camera's image cannot generalize well to another camera's image.

\begin{figure}[tbh]
\centering
\includegraphics[width=0.44\textwidth]{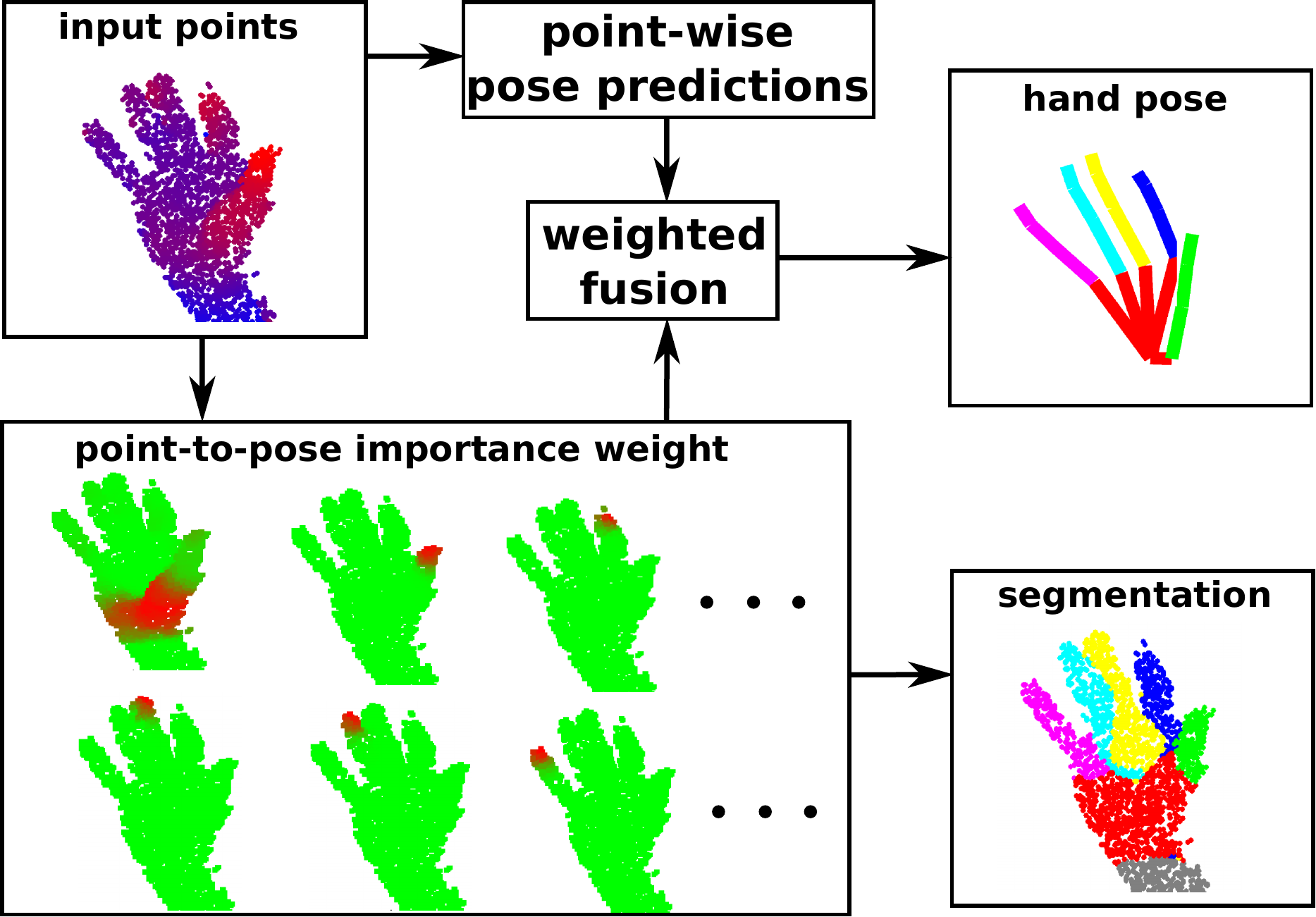}
\caption[teaser] {Our method takes point cloud as input. Then each point predicts the hand pose and its importance weights for different pose dimensions. The final pose is obtained through weighted fusion from each point's pose prediction. Using the importance weight, the hand can be clearly segmented into different parts, although no segmentation ground-truth was used during training.}
\label{fig:teaser}
\end{figure}
Recently, methods using 3D data as input have shown the outperformance over depth image based methods \cite{Yuan2017c}.
One way to use 3D input data is to convert 2D depth image to volumetric representation, such as 3D voxels \cite{Moon2017} \cite{deng2018hand3d}, where occupied 3D voxel is set to 1 and voxels with empty space is set to 0.
Using the voxelized data brings the convenience to directly use 3-dimensional CNN learning structure.
However, the voxelization requires large amount of memory to represent the input and output data, which prevents the deployment of a very deep structure. 

Another way to use 3D input data is to use unordered point cloud as input \cite{ge2018hand}\cite{ge2018point}\cite{chen2018shpr}.
Recently, PointNet, a deep learning structure for point cloud, has shown its success in many tasks.
The PointNet estimates point-wise features for individual points and extract global feature from individual points using a max-pooling layer, such that the network is invariant to the order of points.
Ge et al. use PointNet \cite{qi2017pointnet}\cite{qi2017pointnet++} as backbone to estimate hand pose from point cloud \cite{ge2018hand}. 
However, tedious pre-processing steps such as surface normal estimation and k-nearest-neighours search are required for \cite{ge2018hand}. Moreover, the final max-pooling layer in the PointNet neglects many informations that might be crucial for pose estimation.

\begin{figure*}[t]
\centering
\includegraphics[width=0.99\textwidth]{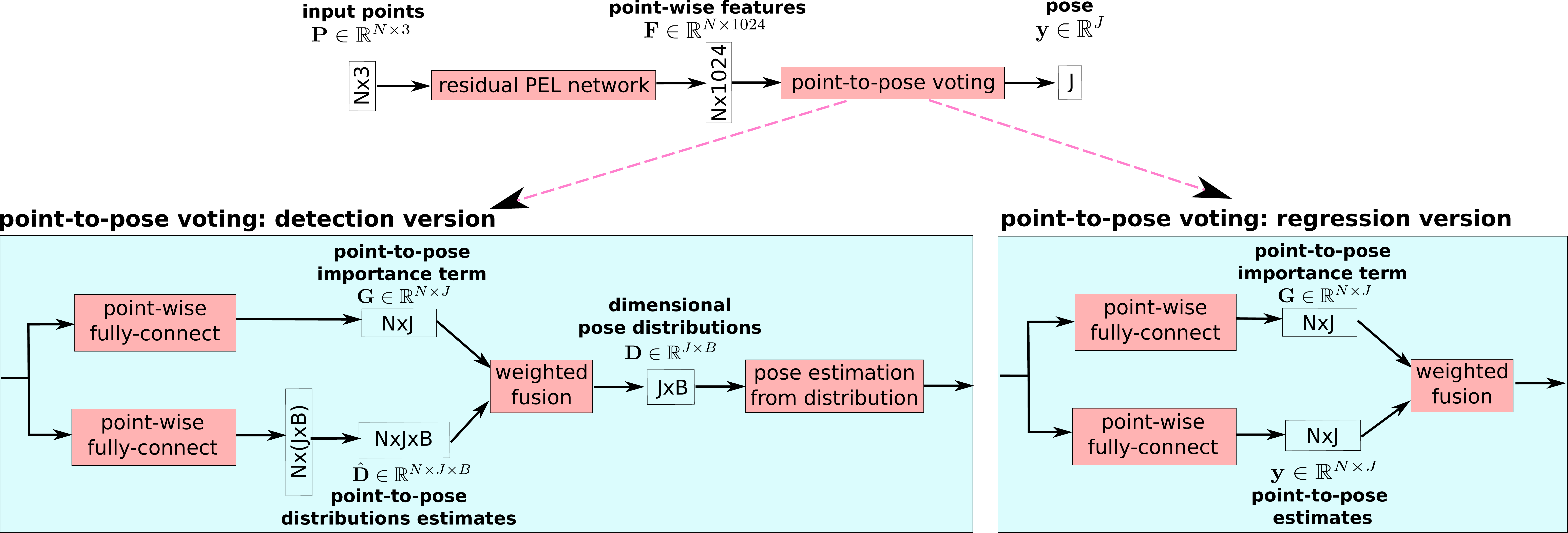}
\caption[overview] {Overview of our method.}
\label{fig:overview}
\end{figure*}

In this work, we explore a more flexible learning structure for unordered point sets, the permutation equivarinat layer (PEL) \cite{ravanbakhsh2016deep} \cite{zaheer2017deep}.
The PEL is a deep learning structure that can be applied for unordered points.
In PEL, point-wise features are computed, where each point's feature dose not only depend on its own input, but also the global maximum value.
Using PEL as the basic element, we propose a residual network version of PEL to construct a deep network for hand pose estimation task.
Moreover, we propose a point-to-pose voting scheme to obtain hand pose, which eliminates the use of max-pooling layer to extract global feature, thus avoiding the loss of information.
Furthermore, the generated point-to-pose importance weights can be also used for the hand segmentation task (Fig. \ref{fig:teaser}), where clear segmentation result can be obtained even without the segmentation ground-truth.

The contributions of this work are:
\begin{itemize}
\item  We propose a novel deep learning based hand pose estimation method for unordered point cloud.
Using Permuation Equivariant Layer as the basic element, a residual network version of PEL is used to solve the hand pose estimation task.
Compared to PointNet \cite{qi2017pointnet++} based methods, our method doesn't require tedious steps such as normal estimation, nearest neighbour estimation.
\item We propose a point-to-pose voting scheme to merge the information from point-wise local features. Furthermore, we show that the voting scheme provides good segmentation results without the need of segmentation ground-truth.
\item We evaluate our method on Hands2017 Challenge dataset and NYU dataset, where state-of-the-art performance is shown. The proposed method achieves the lowest pose error on the Hands2017 Challenge dataset at the time of submission.
\end{itemize}

\section{Related work} 
A lot of research about hand pose estimation has been done in the last decade, which can be categorized to generative, discriminative and hybrid methods. 
Generative methods rely on a hand model and an optimization method to fit the hand model to the observations \cite{romero2017embodied}\cite{tkach2017online}\cite{qian2014realtime}\cite{oikonomidis2011efficient}.
Discriminative methods use learning data to learn a mapping between observation and the hand pose \cite{Oberweger2015a}\cite{Tompson2014}\cite{Moon2017}\cite{deng2018hand3d}\cite{chen2018shpr}\cite{Oberweger2017}\cite{sharp2015accurate}\cite{tang2015opening}.
Hybrid methods use a combination of the generative and discriminative methods \cite{Oberweger2015b} \cite{Sharp2015}\cite{ye2016spatial}.
Our method is a learning based method thus falls into the second category.

\subsection*{Deep learning for hand pose estimation}
With the success of deep learning methods for 2D computer vision, depth image based deep learning methods also showed good performance in hand pose estimation task.
Tompson et al. use 2D CNN to predict heatmaps of each joint and then rely on PSO optimization to estimate the hand pose \cite{Tompson2014}. 
Oberweger et al. \cite{Oberweger2015a} uses 2D CNN to directly regress the hand pose out of the image features, where a bottleneck layer was used to force the predicted pose obey certain prior distribution.
In a later work, Oberweger \cite{Oberweger2017} replaced CNN to a more sophisticated learning structure, ResidualNet50, to improve the performance of feature extraction.
Zhou et al. \cite{Zhou2016} regress a set of hand joint angles and feed the joint angles into an embedded kinematic layer to obtain the final pose.
Ye et al. \cite{ye2018occlusion} use a hierarchical mixture density network to handle the multi-modal distribution of occluded hand joints.

Recently, 3D deep learning has been also applied for the hand pose estimation task.
Moon et al. use $88^3$ voxels to represent hand's 3D geometry and use 3D CNN to estimate hand pose \cite{Moon2017}.
Their method achieved very accurate result, however, 3D voxelization of the input and output data requires large memory size, such that their method only runs at 3.5 FPS.
Ge et al. \cite{ge2018hand}\cite{ge2018point} use 1024 3D points as input, and rely on PointNet \cite{qi2017pointnet++} structure to regress the hand pose. Their method achieved satisfying performance, but tedious pre-processing steps are required, which includes oriented bounding box (OBB) calculation, surface normal estimation and k-nearest-neighbours search for all points.
Chen et al. improves Ge's method by using a spatial transformer network to replace the OBB and furthermore added a auxiliary hand segmentation task to improve the performance \cite{chen2018shpr}. Their method can be trained end-to-end without OBB, but the segmentation ground-truth data require a extra pre-computation step from the pose data.

\subsection*{3D Deep learning}
Since 3D data cannot be directly fed into a conventional 2D CNN, some methods project the 3D data onto different views to obtain multiple depth images and perform CNN on all images \cite{ge2016robust} \cite{qi2016volumetric} \cite{He_2018_CVPR}  \cite{Yu_2018_CVPR}.
Another way to process the 3D data is to use volumetric representation and process the data with 3D CNN \cite{Ge2017}  \cite{wu20153d} \cite{maturana2015voxnet}  \cite{Moon2017}. These methods can capture the feature of input data more effective, but they require large memory size.
Qi et al. developed PointNet to handle unordered point cloud \cite{qi2017pointnet}. The PointNet estimates point-wise local features and obtains global features with a max-pooling layer. Later on, PointNet++ extends PointNet by hierarchically upsampling the local features into higher levels \cite{qi2017pointnet++}.

Other recent methods taking 3D points as input include point-wise CNN \cite{Hua_2018_CVPR}, Deep kd-Networks \cite{klokov2017escape}, Self-Organizing Net \cite{Li_2018_CVPR} and Dynamic Graph CNN \cite{wang2018dynamic}. Despite their good performance for different tasks, they all require extra steps to estimate k-nearest neighbours or construct kd-tree, which are not required in our proposed residual PEL network.

\section{Methodology}
The overview of our method is illustrated in Fig. \ref{fig:overview}. 
Our method takes $N$ 3D points $\mathbf{P}\in\mathbb{R}^{N\times3}$ with arbitrary order as input, and outputs the vectorized 3D hand pose $\mathbf{y}\in\mathbb{R}^J$ in the end, where $J=3\times \#joints$.
To estimate the hand pose, the residual permutation equivariant layers (PEL) (Fig. \ref{fig:respel}) first extract features from each point (Section \ref{subsec:pel}).
Using the point-wise local features, we use point-to-pose voting to estimate the final pose output (Section \ref{subsec:p2p}), where two versions for point-to-pose voting are developed, which are the detection version and the regression version.

\subsection{Pre-processing with view normalization}
%
%
%

\begin{figure}[bth]
\centering
\includegraphics[width=0.47\textwidth]{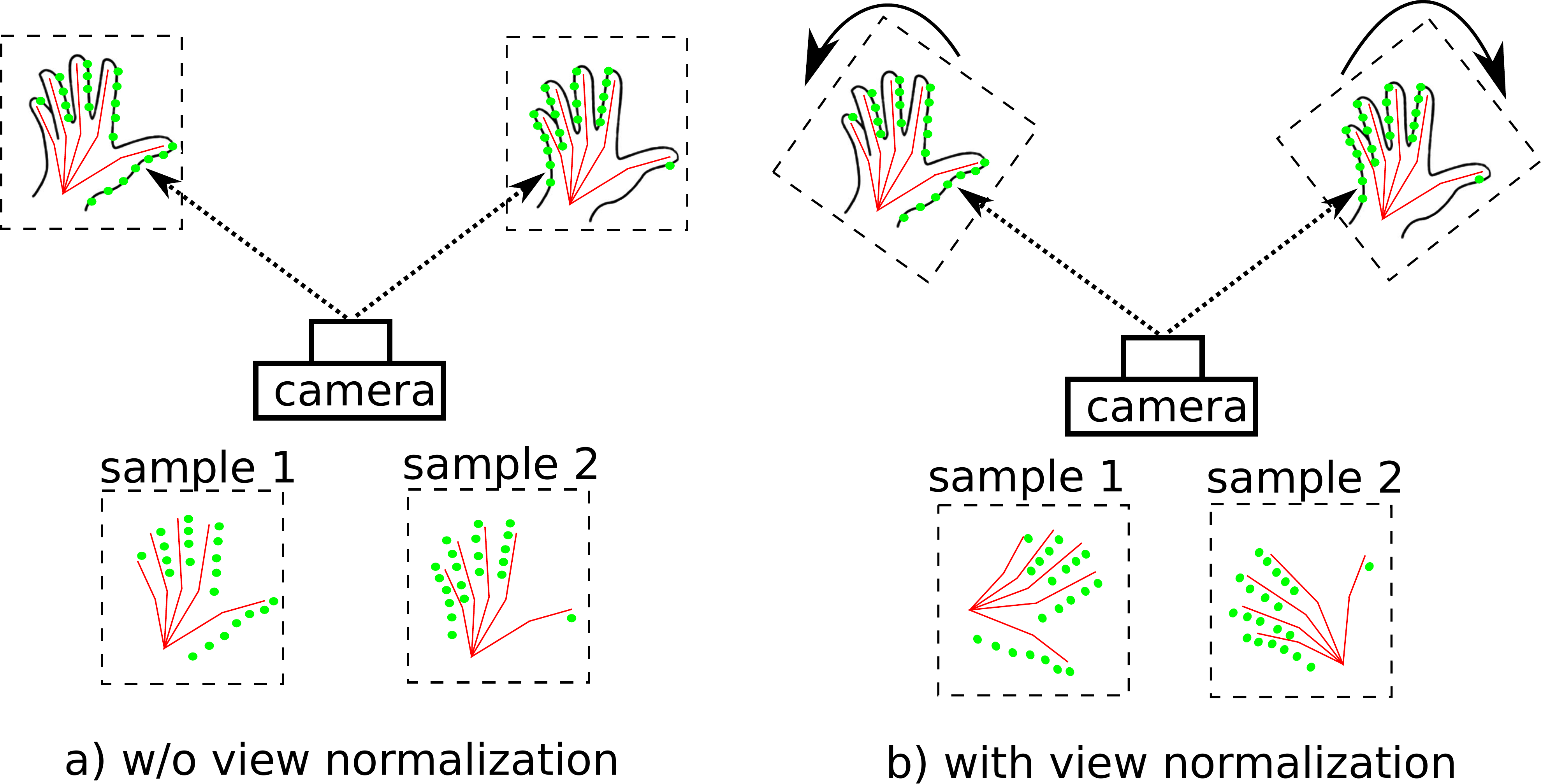}
\caption[viewnorm] {View normalization as pre-processing step. Red skeletons indicate ground-truth pose, green points indicate observed points of the camera. 
a) The same hand pose result in different observations due to different view directions, thus the resulted training samples will contain one-to-many mappings. b) With view normalization, the different observations will also have different pose labels, thus the input-output pairs will have a one-to-one mapping.
}
\label{fig:viewnorm}
\end{figure}

For pre-processing, first, the depth pixels in the hand region are converted to 3D points.
The next step is to create a 3D bounding box for the hand points to obtain normalized coordinate of these points.
A usual pre-processing method will simply create a bounding box aligned with the camera coordinate system (Fig \ref{fig:viewnorm}a).
However, because of self-occlusion of the hand, this will result in different set of observation points for the exact same pose label, which creates one-to-many mapping of the input-output pairs.

To maintain the one-to-one mapping relation of the input-output pairs, we propose to use view normalization to align the bounding box's z-axis $[0,0,1]^T$ with the view direction towards the hand centroid point $\mathbf{c} \in \mathbb{R}^3$.
The alignment is performed by rotating the hand points with a rotation matrix $\mathbf{R}_{cam}$:
\begin{equation} \label{eq:viewnorm}
\begin{split}
\alpha_y &= atan2 (\mathbf{c}_x,\mathbf{c}_z),\\
\widetilde{\mathbf{c}} &= \mathbf{R}_y(-\alpha_y)\cdot\mathbf{c},\\
\alpha_x &= atan2 (\widetilde{\mathbf{c}}_y,\widetilde{\mathbf{c}}_z),\\
\mathbf{R}_{cam} &= \mathbf{R}_y(-\alpha_y)\cdot\boldsymbol{R}_x(\alpha_x).
\end{split}
\end{equation}
After rotating the observation points and ground truth pose with $\mathbf{R}_{cam}$, the hand is rotated such that it appears right in front of the camera,  
As illustrated in Fig. \ref{fig:viewnorm}b, the one-to-many mapping problem is then avoided.

\subsection{Residual Permutation Equivariant Layers}
\label{subsec:pel}
The feature extraction module in our method is called Residual Permutation Equivariant Layers.
The basic element is the permutation equviariant layer (PEL), follows the design from \cite{ravanbakhsh2016deep}.
A PEL takes a set of unordered points as input and computes separate features for each individual input point.

Assuming that the input for a PEL is $\mathbf{x} \in \mathbb{R}^{N\times K_{in}}$ and the output is $\mathbf{x}' \in \mathbb{R}^{NxK_{out}}$, where $N$ is the number of points and $K_{in}, K_{out}$ are the size of input and output feature dimensions.
The output $\mathbf{x}'$ of the PEL is:
\begin{equation}
\mathbf{x}'= \sigma(\boldsymbol{\beta} + (\mathbf{x}\mathbf{I}_{K_{in}}\boldsymbol{\Lambda} + \mathbf{I}_{K_{in}}\mathbf{x}_{max}\boldsymbol{\Gamma})\mathbf{W}),
\end{equation}
where $\Lambda\in\mathbb{R}^{K_{in}}$,$\Gamma\in\mathbb{R}^{K_{in}}$ are weighting terms for the point's own feature and the global maximum value respectively, and $\mathbf{x}_{max}\in{K_{in}}$ is a vector representing maximum values for each column of $\mathbf{x}$. 
$\mathbf{W}\in\mathbb{R}^{K_{in} \times K_{out}}$ is the weight term and 
$\boldsymbol{\beta}\in\mathbb{R}^{k_{out}}$ is the bias term. 
Furthermore, an activation function $\sigma(\cdot)$ is applied to provide non-linearity, where a sigmoid function is used in our method.

This layer is invariant to input order because the output value of each individual point only depend on its own input feature and the global maximum values in each feature dimension, whereas the global maximum values are also invariant to the order of input points. 
\footnote{The detailed proof of the invariance for PEL can be found in \cite{ravanbakhsh2016deep}.}
In this way, each point's feature is not only computed based on its own input feature, each point also exchanges information with other points through the weighted summation of $\mathbf{x}_{max}$.

For the practical side, four elements need to be trained, which are $\mathbf{\beta}$, $\boldsymbol{\Lambda}$, $\boldsymbol{\Gamma}$ and $\mathbf{W}$. 
In total, the number of parameters needed for one layer is $K_{out} + (K_{out}+2)K_{in}$, which is only slightly more than a fully-connected layer, thus it is feasible for training in practice.

\begin{figure}[h]
\centering
\includegraphics[width=0.4\textwidth]{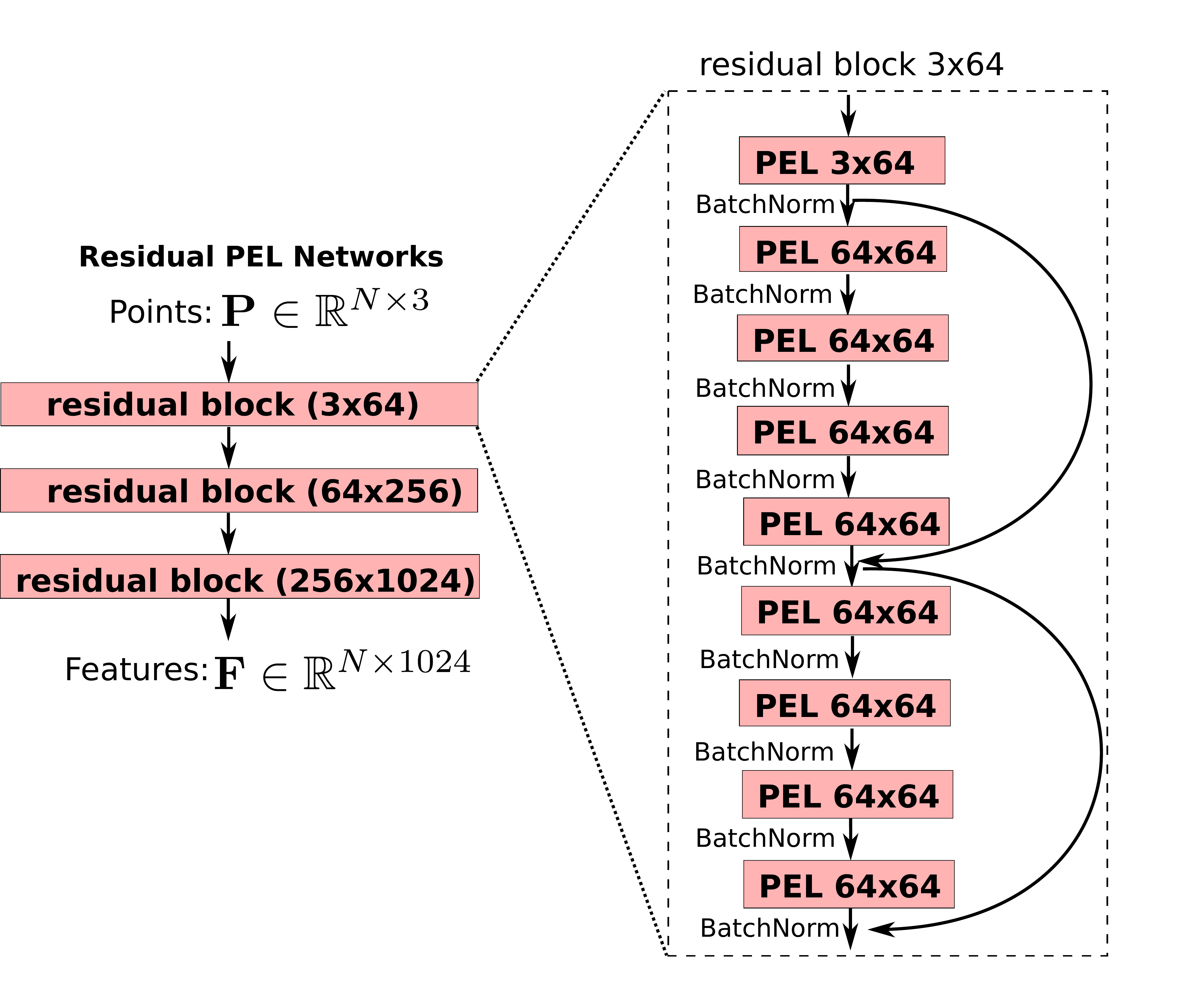} 
\caption[respel] {Residual network of permutation equivariant layer}
\label{fig:respel}
\end{figure}

In order to extract very complex features, we construct a residual network with 27 PEL layers.
As illustrated in Fig. \ref{fig:respel}, we use three residual blocks, whereas each residual block consists of 9 PELs and three short-cut connections.
Furthermore, after each PEL, a batch normalization is performed.

\subsection{Point-to-pose voting}
\label{subsec:p2p}
With the residual PEL module, features $\mathbf{F}$ of points are computed, where each row of $\mathbf{F}$ represents local feature for one point.
Using these local point-wise features, the hand pose $\mathbf{y}\in \mathbb{R}^{J}$ will be estimated using a point-to-pose voting scheme.
Two versions for point-to-pose voting are explored, which are the detection based version and the regression based version.
The performance of these two versions will be compared in the experiment section.
\subsubsection*{Detection version}
In the detection version (Fig. \ref{fig:overview} left), probability distributions of each pose dimension is firstly detected and the pose is then integrated from the distributions.
We use two separate fully connected modules to estimate two matrices: an importance term $\mathbf{G}\in\mathbb{R}^{N\times J}$ and a distributions $\hat{\mathbf{D}}\in\mathbb{R}^{N\times J \times B}$.
An element of importance matrix $\mathbf{G}_{nj}$ represents the confidence level of $n$th input point to predict the $j$th output pose dimension.
In other words, each of the $N$ points predicts $J$ $B$-dimensional distributions and $J$ corresponding importance weights.
Notice that the final layer of the two fully connected modules are sigmoid functions, such that all elements of $\mathbf{G}$ and $\hat{\mathbf{D}}$ are in the range of $[0,1]$.

$\hat{\mathbf{D}}$ represents the output pose distributions, where each point makes its own predictions to $J$ output dimensions.
Each of the output pose dimension is represented as discrete distribution using a $B$ bins, representing the value range in $[-r, +r]$ with the resolution per bin $\Delta d = 2r/B$.
For the $j$th dimension of the output pose $\mathbf{y}_{j}$, the corresponding bin index for itself is then: 
$$index^{gt}_j = \ceil{(\mathbf{y}_{j}^{gt}+r)/\Delta d},$$
and the ground truth distribution is defined as:
\begin{equation} 
\mathbf{D}_{jb}^{gt} = 
\begin{cases}
      1, & \text{if}\  b \in [index^{gt}_j-1, index^{gt}_j+1]  \\
      0, & \text{otherwise}
\end{cases}
\end{equation}
whereas the three bins around ground truth pose are set to one and all other bins are set to zero.
%

The final distribution for the $J$-dimensional output $\mathbf{D} \in\mathbb{R}^{J\times B}$ is then obtained by merging the predictions of all $N$ points: 
\begin{equation}
\mathbf{D}_{jb} = \frac{\sum_{n=1}^{N} (\mathbf{G}_{nj} \hat{\mathbf{D}}_{njb})}{\sum_{n=1}^{N}\mathbf{G}_{nj}} .
\end{equation}

And the final pose $\mathbf{y}$ is estimated with integration over the distribution:
\begin{equation}
\mathbf{y}_j = \frac{\sum_{b=1}^{B} (b-0.5)\mathbf{D}_{jb}}{\sum_{b=1}^{B} \mathbf{D}_{jb}},
\end{equation}
where $b-0.5$ represents the bin center position.


\begin{table*}[thb]
\label{tb:selfcomp}
\centering
\small
\begin{tabular}{c|ccc|ccc}
 & \multicolumn{3}{c|}{Hands2017Challenge dataset} & \multicolumn{2}{c}{NYU dataset} \\ \hline
 & \multicolumn{1}{c}{detection} & \multicolumn{1}{c}{\begin{tabular}[c]{@{}c@{}}detection w/o\\ view normalization\end{tabular}} & \multicolumn{1}{c|}{regression} & \multicolumn{1}{c}{\begin{tabular}[c]{@{}c@{}}detection/\\ single view\end{tabular}}  & \multicolumn{1}{c}{\begin{tabular}[c]{@{}c@{}}regression/\\ single view\end{tabular}}   & \multicolumn{1}{c}{\begin{tabular}[c]{@{}c@{}}regression/\\ three views\end{tabular}} \\ \hline
256 points & 11.34 &13.14 & 11.21 & 9.82 & 9.45 & 9.05\\
512 points & 10.23 & 11.93 & 10.11 & 9.33  & 9.06 & 8.49 \\
1024 points &  9.93& 11.67  & 9.82 & 9.25 & 8.99 & 8.35\\
2048 points & 9.93 & 11.69 & 9.87 & 9.32 &  9.08 & 8.35
\end{tabular}
\caption{Self-comparison result}
\end{table*}

\subsubsection*{Regression version}
In the regression version (Fig. \ref{fig:overview} right), each point will directly predict the pose without the intermediate distribution detection. 
Similarly to the detection version, two separate fully connected modules are used to estimate the importance term $\mathbf{G}\in\mathbb{R}^{N\times J}$ and the point-to-pose estimates $\hat{\mathbf{y}}\in\mathbb{R}^{N\times J}$.
Then the final pose output is merged as the weighted average over all points' predictions:
\begin{equation} 
\mathbf{y}_{j} = \frac{\sum_{n=1}^{N} (\mathbf{G}_{nj} \hat{\mathbf{y}}_{nj})}{\sum_{n=1}^{N}\mathbf{G}_{nj}}.
\end{equation}

\subsection{Segmentation using importance term}
The importance term $\mathbf{G}\in\mathbb{R}^{N\times J}$ is estimated automatically without the ground-truth information.
However, it still provides vital information of each point's importance to the pose output.
Therefore, the obtained importance term can be also used for the hand segmentation task based on the most contributed pose dimension.
For the $n$-th point having the importance terms $\mathbf{g} = \mathbf{G}_n$, the point's most contributed pose dimension is:
$$j_{max} = \argmax_j \mathbf{g}_j,$$
where the pose dimension $j_{max}$ can be categorized to a specific hand part.
In this work, we categorized the $J$ pose dimensions to palm, thumb, index, ring and pinky fingers.
 
\begin{figure*}[htb]
\centering
\includegraphics[width=0.999\textwidth]{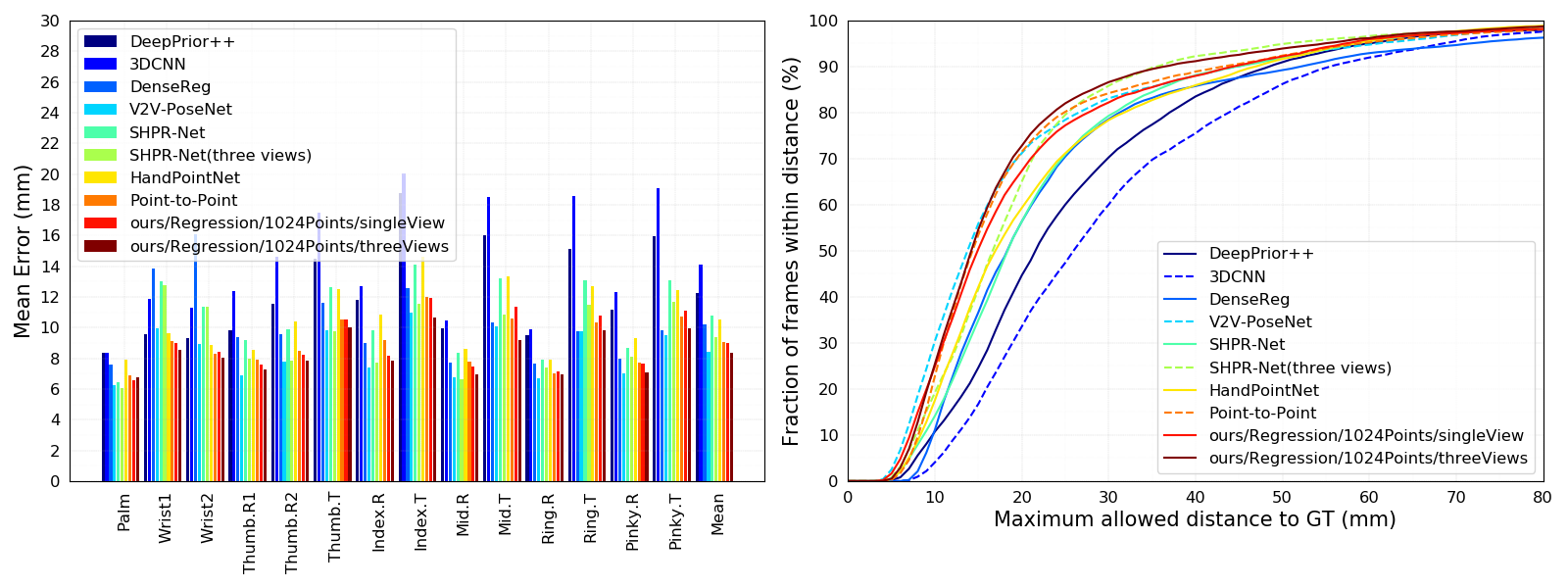}
\caption[nyucompare] {Comparison with state-of-the-arts on NYU \cite{Tompson2014} dataset. Left: mean errors of different joints. Right: proportion of correct frames based on different error thresholds.}
\label{fig:nyucompare}
\vspace{-0.4cm}
\end{figure*}

\begin{figure*}[htb]
\centering
\includegraphics[width=0.97\textwidth]{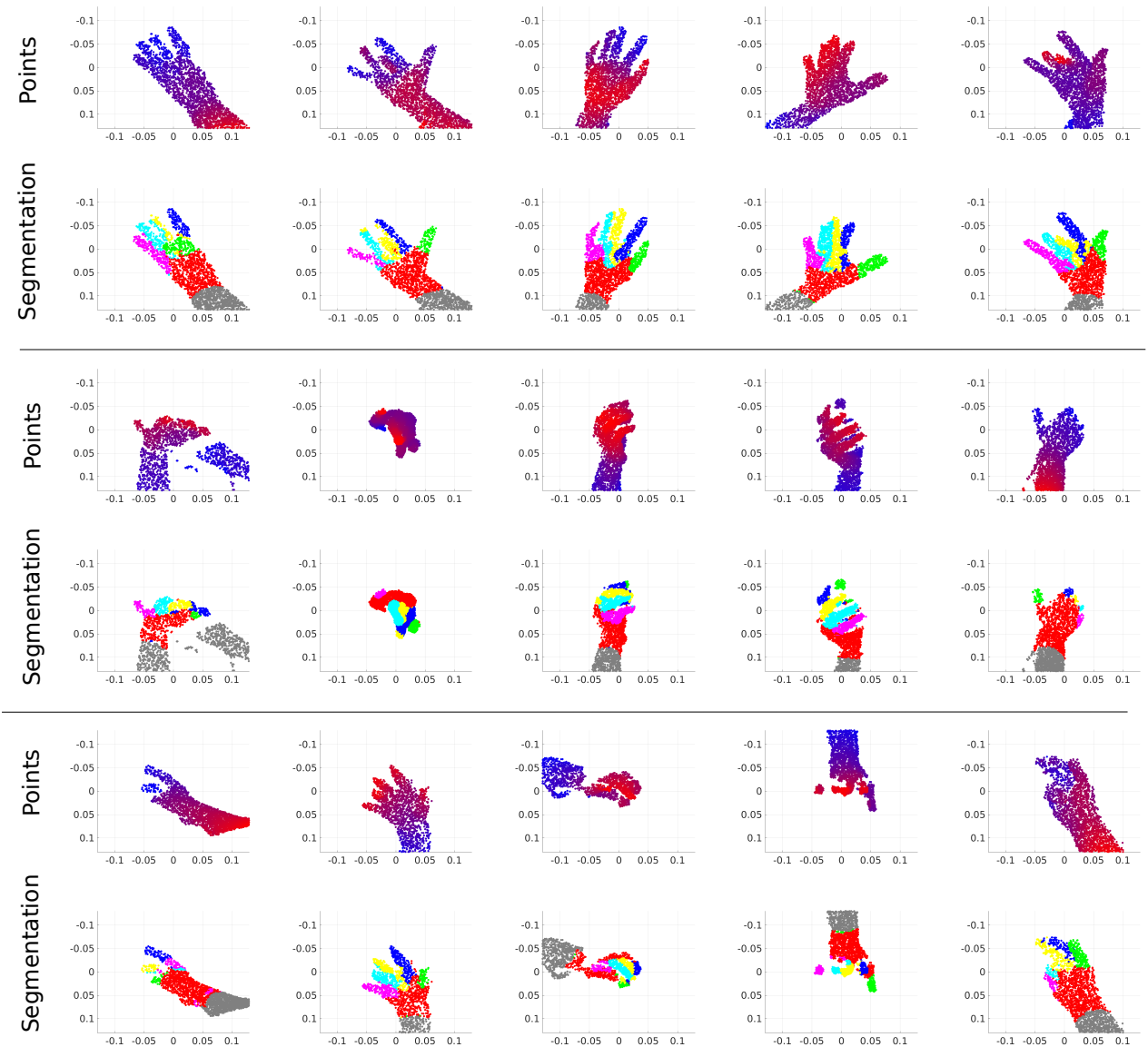}
\caption[segmentation] {Segmentation results based on importance weights (best viewed in color). Points: input point cloud, color indicates depth value, blue points are more distanced and red points are more closer to the camera. Segmentation: each part of the hand is indicated with an different color, palm (red), thumb (green), index (blue), middle (yellow), ring (cyan), pinky (pink) and irrelevant points with low importance weight for all parts (gray)}
\label{fig:seg}
\end{figure*}

\subsection{Training Loss}
The only training loss for the detection version is the logarithm loss of the pose distributions:
\begin{equation}
\begin{split}
L_{det} = -\sum_{j=1}^J \sum_{b=1}^B &\mathbf{D}_{jb}^{gt}  log(\mathbf{D}_{jb} +\epsilon) \\
+ &(1-\mathbf{D}_{jb}^{gt}) (1- log(\mathbf{D}_{jb} +\epsilon)) ,
\end{split}
\end{equation}
where $\epsilon=10^{-7}$ is a small offset to avoid feeding zero to the logarithm operator.

The only training loss used for the regression version is the L2 loss between predicted pose and ground-truth pose:
\begin{equation}
L_{reg} = \frac{1}{2} \sum_{j=1}^J   (\mathbf{y}_{j}^{gt} -   \mathbf{y}_{j})^2.
\end{equation}

For both detection and regression versions, the importance term $\mathbf{G}\in\mathbb{R}^{N\times J}$ is estimated automatically without the ground-truth information.

\section{Experiment and result}

Our hand pose estimation method is evaluated on the Hands2017Challenge dataset \cite{Yuan2017b} and the NYU \cite{Tompson2014} dataset. 
The Hands2017Challenge is composed from parts of the Big Hand 2.2M dataset \cite{Yuan2017a} and the First-person Hand Action Dataset (FHAD) \cite{Garcia-Hernando2017}, it is currently the largest dataset available. 
Its training set contains 957032 depth images of five different hands. The test set consists of 295510 depth images of ten different hand shapes, of which five are the same as in the training set and five are entirely new. 
The NYU dataset contains 72757 training images of a single subject's hand and 8252 test images that include a second hand shape besides the one from the training set.
The NYU dataset provides depth images from three different views, we trained our method both using only frontal view data and using all three views. And we test only using the frontal view.

Our method is implemented using TensorFlow \cite{Abadi2016}.
The networks are trained on a PC with an AMD FX-4300/Intel Core i7-860 CPU and an nVidia GeForce GTX1060 6GB GPU. 
We train 100 epochs for the NYU dataset and train only 20 epochs for the Hands 2017 challenge dataset since the challenge dataset has a large size.
For both datasets, the first 50\% of the epochs are trained with smaller point size ($N=256$) to boost the training speed.
The remaining epochs are trained with a point size of $N=512$.
We used Adam optimizer for training with an initial learning rate of $10^{-3}$ and we decrease the learning rate to $10^{-4}$ for the last 10\% of the epochs.
For the detection version, we set $r=15mm$ and $B=60$.
Online augmentation was performed with random translation in all three dimensions within $[-15, 15] mm$, random scaling within $[0.85,1.15]$ and random rotation around z-axis within $[-\pi,\pi]$.

\subsection{Evaluation metrics}
For the NYU dataset, two standard metrics are used to evaluated the performance.
The first metric is the mean joint error, which measures the average Euclidean distance error for all joints across the whole test set.
The second metric is correct frame proportion, which indicates the proportion of frames that have all joints within a certain distance to ground truth.
The second metric is considered as more difficult since single joint violation will cause an unqualified frame.
For the Hands2017Challenge dataset, only the mean joint error is used since the ground-truth data of test set is not publicly available and the official test website only provides the mean joint error result.

\begin{table}[tbh]
\centering
\small
\begin{tabular}{c|ccc}
\textbf{method} & \multicolumn{1}{l}{\textbf{avg test}} & \multicolumn{1}{l}{\textbf{seen test}} & \multicolumn{1}{l}{\textbf{unseen test}} \\ \hline
Ours/regression & 9.82                      &7.15                       & 12.04                                 \\
Ours/detection   & 9.93                      &7.18                       &  12.22                               \\ \hline
V2V-PoseNet \cite{Moon2017}     & 9.95                      & 6.97                      & 12.43                     \\
RCN-3D \cite{Yuan2017c}           & 9.97                      & 7.55                      & 12.00                     \\
oasis   \cite{ge2018hand}              & 11.30                    & 8.86                      & 13.33                      \\
THU VCLab   \cite{chen2017pose}    & 11.70                    & 9.15                      & 13.83                           \\
Vanora   \cite{Yuan2017c}           & 11.91                    & 9.55                      & 13.89                          
\end{tabular}
\caption{Comparison of our method with state-of-the-art methods on the Hands2017Challenge dataset}
\vspace{-0.5cm}
\end{table}


\begin{table}[thb]
\centering
\small
\begin{tabular}{c|ccc}
\textbf{method} & \multicolumn{1}{l}{\textbf{mean joint error (mm)}} \\ \hline
Ours/regression/singleView & 8.99                                                  \\
Ours/regression/threeViews  & 8.35                                  \\ 
Ours/detection   &   9.25   \\  \hline
DeepPrior++ \cite{Oberweger2017}  & 12.23                        \\
3DCNN         & 14.11                             \\
DenseReg   \cite{deng2018hand3d}              & 10.21                                   \\
V2V-PoseNet  \cite{Moon2017}    & 8.42                           \\
SHPR-Net  \cite{chen2018shpr}           & 10.77                                \\
SHPR-Net (three views) \cite{chen2018shpr}           & 9.37                                \\

HandPointNet  \cite{ge2018hand}    &  10.54\\
Point-to-Point \cite{ge2018point}     &  9.04\\
\end{tabular}
\caption{Comparison of our method with state-of-the-art methods on the NYU dataset}
\vspace{-0.5cm}
\end{table}

\subsection{Self-comparison}
\vspace{-0.1cm}

In this subsection, we perform self-comparison to show the effects of different components in our method.
The detailed comparison can be found in Table 1.

\textbf{View normalization.}
To validate the necessity of view normalization, we trained our method using both view normalized data and original data for the detection version.
It is evident from Table 1 that view normalization decreases the pose estimation error by about 1.5 mm for the Hands2017Challenge dataset.
\vspace{-0.05cm}

\textbf{Detection vs. regression.}
Yuan et.al. indicates that detection based methods work in general better than regression-based methods \cite{Yuan2017c}, therefore we implemented both detection-based (ours/distribution) and regression-based (ours/regression) variations.
As seen from Table 1, in both datasets, both variations show similar performance, where regression-based variation slightly outperforms the detection-based counterpart.
Possible reasons for this can be quantization effect of the binary distribution and the simplification of 1-dimensional heat vectors compared to 2D or 3D heat maps used in previous works.
However, the 1D heat vector representation is much more efficient than the 3D heatmap representation.
For the heat vectors, we need $B\times J$ values to represent the pose output, whereas 3D heatmaps require $J\times B^3$ values \cite{Moon2017}. 
In future work, it is worth to investigate more different loss types and heat map representations.

\vspace{-0.05cm}
\textbf{Number of points.}
Taking advantage of the PEL structure and voting-based scheme, our method is very flexible to the input point cloud size.
Although the network was trained with 512 points, arbitrary number of points can be used at the testing stage.
For an online application, this property can be beneficial to choose an arbitrary number of points based on the computational resources available.
As seen from Table 1, different point sizes were tested for both datasets.
Our method can achieve good performance with only 256 points, the mean joint error only increased by 0.11 mm compared to 512 points.
In general, more points provides better performance, but it doesn't improve any more after 1024 points. 
Therefore, we choose 1024 points for testing to compare our method with other state-of-the-art methods.

\begin{table*}[thb]
\label{tb:runtime}
\centering
\small
\begin{tabular}{c|c|c|c|c|c|c}
\multicolumn{1}{l|}{}    & \multicolumn{3}{c|}{our method}         & V2V-PoseNet  \cite{Moon2017}      & Hand3D    \cite{deng2018hand3d}               & P2P-Regression \cite{ge2018point}         \\ \hline
\multicolumn{1}{l|}{GPU} & \multicolumn{3}{c|}{GTX1060}            & Titan X                   & Titan X                  & Titan Xp                 \\ \hline
\multirow{4}{*}{time}    &             & detection & regression & \multirow{4}{*}{285.7 ms} & \multirow{4}{*}{33.3 ms} & \multirow{4}{*}{23.9 ms} \\ \cline{2-4}
                         & 256 points  & 3.5 ms       & 2.9 ms     &                           &                          &                          \\
                         & 512 points  & 6.9 ms       & 5.5 ms     &                           &                          &                          \\
                         & 1024 points & 12.5 ms      & 10.7 ms    &                           &                          &                         
\end{tabular}
\caption{Comparison of runtime and hardware}
\end{table*}

\subsection{Comparison to state-of-the-art methods}
\textbf{Hands2017Challenge dataset}.
Since the ground-truth data for the testing set are not publicly available, some previous papers divide the training set on their own to create their own testing set.
Therefore, for fair comparison, we only compare to those methods, who have also tested on the official testing website\footnote{https://competitions.codalab.org/competitions/17356\#learn\_the\_de}. 
In Table 2, we compare our method with five other top performing methods on the Hands2017Challenge dataset, which include both methods using 3D input data and methods using 2D depth image.
RCN-3D \cite{Yuan2017c}, THU VCLab  \cite{chen2017pose} and Vanora \cite{Yuan2017c} use depth image as input data.
V2V-PoseNet  \cite{Moon2017}  uses voxel representation for both input data and output heatmaps. 
Oasis \cite{ge2018hand} also uses 3D point cloud as input and their method is constructed based on PointNet \cite{qi2017pointnet}. 
Three different errors are used for comparison: 1) the average across the complete test set (\textit{avg test}), 2) the average across the test set of seen subjects' hand during training (\textit{seen test}), and 3) the average across the test set images of unseen subjects' hand (\textit{unseen test}).
Currently, our method achieves the lowest overall mean joint error on the test dataset of 9.82 mm.
For seen subjects' hand and unseen subjects' hand, the mean joint errors are 7.15 mm and 12.04 mm respectively, which shows the generalizability of the proposed method even without regularization on the parameters.
In comparison to other 3D data based methods, our method is slightly better than V2V-PoseNet, whereas V2V-PoseNet requires 10 good GPUs to run realtime and our method requires only one moderate GPU.
Compared to oasis, which also uses 1024 3D points as input, our method is 1.48 mm better, where oasis requires more input information like surface normal and k-nearest neighbours.

\textbf{NYU dataset}.
For the NYU dataset, we only compared to recent state-of-the-art methods after 2017.
For both training and testing, only the frontal view was used.
Following previous works \cite{Oberweger2015a}\cite{Tompson2014}\cite{ge2018point}, only 14 joints  out of 36 joints provided were used for evaluation.
For a fair comparison, we only compared to the methods trained solely on the NYU dataset without additional data.
The compared methods include depth image based methods (DeepPrior++ \cite{Oberweger2017}, DenseReg \cite{deng2018hand3d}), 3D voxel based methods (3DCNN \cite{Ge2017}, V2V-PoseNet \cite{Moon2017}) and point cloud based methods (SHPR-Net \cite{chen2018shpr}, HandPointNet \cite{ge2018hand}, Point-to-Point \cite{ge2018point}).
The comparison is shown in Figure \ref{fig:nyucompare}, where our method performs comparably good with V2V-PoseNet \cite{Moon2017} and Point-to-Point \cite{ge2018point}, and outperforms all other methods.
A closer comparison of the mean joint error value can be found in Table 3, where our method using single view is the second best, and our method with three views outperforms all recent state-of-the-art methods.


\subsection{Segmentation using importance term}
Besides showing the quantitative results relying on the ground-truth data, we also show some qualitative result of the segmentation using the automatically inferred importance term. 
As seen from Figure \ref{fig:seg}, the segmentation result is shown alongside the original point cloud.
The samples are taken from the Hands2017Challenge dataset. Both samples with all visible fingers and samples with different levels of self-occlusion are shown.
In all cases, the fingers are clearly segmented with each other, even the fingers are twisted together.
The points has no contribution to any joint has very small importance values and they are classified as background. As Figure \ref{fig:seg} shows, the arm and the background points are clearly segmented in gray.
Notice that the segmentation result is obtained without the ground-truth data for segmentation. This leads to a future research question about whether we can perform this method on hand-object interaction cases, where the influence of the object can be automatically removed.

\subsection{Runtime and model size}
To store the learned models, the proposed method takes 38 MB for the regression version and 44 MB for the detection version.
Compared to 420MB for a 3D CNN based method \cite{deng2018hand3d}, our model size is much smaller.
For the testing stage, the runtime of our method is 12.5 ms and 10.7 ms per frame for the detection and regression version respectively, where 1024 points are used as input. 
When less input points is used, the runtime can be further reduced with a small performance loss.
Table 4 shows a comparison of runtime to other state-of-the-art 3D methods \cite{Moon2017}\cite{deng2018hand3d}\cite{ge2018point}  . 
Although the other methods all used a more powerful GPU than ours, our method require the least processing time.

\section{Conclusion}
We propose a novel neural network architecture, ResidualPEL, for hand pose estimation using unordered point cloud as input.
The proposed method is invariant to input point order and can handle different numbers of points.
Compared to previous 3D voxel based methods, our method requires less memory size. And compared to PointNet based methods, our method does not require surface normal and k-nearest-neighours information.
A voting-based scheme was proposed to merge information from individual points to pose output, where the resulting importance term can be also used to segment the hand into different parts.
The performance of our method is evaluated on two datasets, where our method outperforms the state-of-the-art methods on both datasets.
In future work, the proposed ResidualPEL and voting scheme can be also applied to similar problem such as human pose estimation and object pose estimation.
{\small
\bibliographystyle{ieee}
\bibliography{egbib}

\begin{thebibliography}{10}\itemsep=-1pt

\bibitem{Abadi2016}
M.~Abadi, A.~Agarwal, P.~Barham, E.~Brevdo, Z.~Chen, C.~Citro, G.~S. Corrado,
  A.~Davis, J.~Dean, M.~Devin, et~al.
\newblock Tensorflow: Large-scale machine learning on heterogeneous distributed
  systems.
\newblock {\em arXiv:1603.04467}, 2016.

\bibitem{chen2017pose}
X.~Chen, G.~Wang, H.~Guo, and C.~Zhang.
\newblock Pose guided structured region ensemble network for cascaded hand pose
  estimation.
\newblock {\em arXiv preprint arXiv:1708.03416}, 2017.

\bibitem{chen2018shpr}
X.~Chen, G.~Wang, C.~Zhang, T.-K. Kim, and X.~Ji.
\newblock Shpr-net: Deep semantic hand pose regression from point clouds.
\newblock {\em IEEE Access}, 6:43425--43439, 2018.

\bibitem{deng2018hand3d}
X.~Deng, S.~Yang, Y.~Zhang, P.~Tan, L.~Chang, and H.~Wang.
\newblock Hand3d: Hand pose estimation using 3d neural network.
\newblock {\em Proc. Computer Vision and Pattern Recognition (CVPR), IEEE},
  2018.

\bibitem{Garcia-Hernando2017}
G.~Garcia-Hernando, S.~Yuan, S.~Baek, and T.-K. Kim.
\newblock First-person hand action benchmark with rgb-d videos and 3d hand pose
  annotations.
\newblock {\em arXiv:1704.02463}, 2017.

\bibitem{ge2018hand}
L.~Ge, Y.~Cai, J.~Weng, and J.~Yuan.
\newblock Hand pointnet: 3d hand pose estimation using point sets.
\newblock In {\em Proceedings of the IEEE Conference on Computer Vision and
  Pattern Recognition}, pages 8417--8426, 2018.

\bibitem{ge2016robust}
L.~Ge, H.~Liang, J.~Yuan, and D.~Thalmann.
\newblock Robust 3d hand pose estimation in single depth images: from
  single-view cnn to multi-view cnns.
\newblock In {\em Proceedings of the IEEE conference on computer vision and
  pattern recognition}, pages 3593--3601, 2016.

\bibitem{Ge2017}
L.~Ge, H.~Liang, J.~Yuan, and D.~Thalmann.
\newblock 3d convolutional neural networks for efficient and robust hand pose
  estimation from single depth images.
\newblock In {\em IEEE conference on computer vision and pattern recognition},
  pages 1991--2000, 2017.

\bibitem{ge2018point}
L.~Ge, Z.~Ren, and J.~Yuan.
\newblock Point-to-point regression pointnet for 3d hand pose estimation.
\newblock {\em ECCV, Springer}, 1, 2018.

\bibitem{He_2018_CVPR}
X.~He, Y.~Zhou, Z.~Zhou, S.~Bai, and X.~Bai.
\newblock Triplet-center loss for multi-view 3d object retrieval.
\newblock In {\em The IEEE Conference on Computer Vision and Pattern
  Recognition (CVPR)}, June 2018.

\bibitem{Hua_2018_CVPR}
B.-S. Hua, M.-K. Tran, and S.-K. Yeung.
\newblock Pointwise convolutional neural networks.
\newblock In {\em The IEEE Conference on Computer Vision and Pattern
  Recognition (CVPR)}, June 2018.

\bibitem{klokov2017escape}
R.~Klokov and V.~Lempitsky.
\newblock Escape from cells: Deep kd-networks for the recognition of 3d point
  cloud models.
\newblock In {\em Computer Vision (ICCV), 2017 IEEE International Conference
  on}, pages 863--872. IEEE, 2017.

\bibitem{Li_2018_CVPR}
J.~Li, B.~M. Chen, and G.~Hee~Lee.
\newblock So-net: Self-organizing network for point cloud analysis.
\newblock In {\em The IEEE Conference on Computer Vision and Pattern
  Recognition (CVPR)}, June 2018.

\bibitem{maturana2015voxnet}
D.~Maturana and S.~Scherer.
\newblock Voxnet: A 3d convolutional neural network for real-time object
  recognition.
\newblock In {\em Intelligent Robots and Systems (IROS), 2015 IEEE/RSJ
  International Conference on}, pages 922--928. IEEE, 2015.

\bibitem{Moon2017}
G.~Moon, J.~Y. Chang, and K.~M. Lee.
\newblock V2v-posenet: Voxel-to-voxel prediction network for accurate 3d hand
  and human pose estimation from a single depth map.
\newblock {\em arXiv:1711.07399}, 2017.

\bibitem{Oberweger2017}
M.~Oberweger and V.~Lepetit.
\newblock Deepprior++: Improving fast and accurate 3d hand pose estimation.
\newblock In {\em ICCV workshop}, volume 840, page~2, 2017.

\bibitem{Oberweger2015a}
M.~Oberweger, P.~Wohlhart, and V.~Lepetit.
\newblock Hands deep in deep learning for hand pose estimation.
\newblock In {\em Computer Vision Winter Workshop}, pages 1--10, 2015.

\bibitem{Oberweger2015b}
M.~Oberweger, P.~Wohlhart, and V.~Lepetit.
\newblock Training a feedback loop for hand pose estimation.
\newblock In {\em IEEE International Conference on Computer Vision}, pages
  3316--3324, 2015.

\bibitem{oikonomidis2011efficient}
I.~Oikonomidis, N.~Kyriazis, and A.~A. Argyros.
\newblock Efficient model-based 3d tracking of hand articulations using kinect.
\newblock In {\em BmVC}, volume~1, page~3, 2011.

\bibitem{qi2017pointnet}
C.~R. Qi, H.~Su, K.~Mo, and L.~J. Guibas.
\newblock Pointnet: Deep learning on point sets for 3d classification and
  segmentation.
\newblock {\em Proc. Computer Vision and Pattern Recognition (CVPR), IEEE},
  1(2):4, 2017.

\bibitem{qi2016volumetric}
C.~R. Qi, H.~Su, M.~Nie{\ss}ner, A.~Dai, M.~Yan, and L.~J. Guibas.
\newblock Volumetric and multi-view cnns for object classification on 3d data.
\newblock In {\em Proceedings of the IEEE conference on computer vision and
  pattern recognition}, pages 5648--5656, 2016.

\bibitem{qi2017pointnet++}
C.~R. Qi, L.~Yi, H.~Su, and L.~J. Guibas.
\newblock Pointnet++: Deep hierarchical feature learning on point sets in a
  metric space.
\newblock In {\em Advances in Neural Information Processing Systems}, pages
  5099--5108, 2017.

\bibitem{qian2014realtime}
C.~Qian, X.~Sun, Y.~Wei, X.~Tang, and J.~Sun.
\newblock Realtime and robust hand tracking from depth.
\newblock In {\em Proceedings of the IEEE conference on computer vision and
  pattern recognition}, pages 1106--1113, 2014.

\bibitem{ravanbakhsh2016deep}
S.~Ravanbakhsh, J.~Schneider, and B.~Poczos.
\newblock Deep learning with sets and point clouds.
\newblock {\em arXiv preprint arXiv:1611.04500}, 2016.

\bibitem{romero2017embodied}
J.~Romero, D.~Tzionas, and M.~J. Black.
\newblock Embodied hands: Modeling and capturing hands and bodies together.
\newblock {\em ACM Transactions on Graphics (TOG)}, 36(6):245, 2017.

\bibitem{sharp2015accurate}
T.~Sharp, C.~Keskin, D.~Robertson, J.~Taylor, J.~Shotton, D.~Kim, C.~Rhemann,
  I.~Leichter, A.~Vinnikov, Y.~Wei, et~al.
\newblock Accurate, robust, and flexible real-time hand tracking.
\newblock In {\em Proceedings of the 33rd Annual ACM Conference on Human
  Factors in Computing Systems}, pages 3633--3642. ACM, 2015.

\bibitem{Sharp2015}
T.~Sharp, C.~Keskin, D.~Robertson, J.~Taylor, J.~Shotton, D.~Kim, C.~Rhemann,
  I.~Leichter, A.~Vinnikov, Y.~Wei, et~al.
\newblock Accurate, robust, and flexible real-time hand tracking.
\newblock In {\em 33rd ACM Conference on Human Factors in Computing Systems},
  pages 3633--3642. ACM, 2015.

\bibitem{tang2015opening}
D.~Tang, J.~Taylor, P.~Kohli, C.~Keskin, T.-K. Kim, and J.~Shotton.
\newblock Opening the black box: Hierarchical sampling optimization for
  estimating human hand pose.
\newblock In {\em Proceedings of the IEEE international conference on computer
  vision}, pages 3325--3333, 2015.

\bibitem{tkach2017online}
A.~Tkach, A.~Tagliasacchi, E.~Remelli, M.~Pauly, and A.~Fitzgibbon.
\newblock Online generative model personalization for hand tracking.
\newblock {\em ACM Transactions on Graphics (TOG)}, 36(6):243, 2017.

\bibitem{Tompson2014}
J.~Tompson, M.~Stein, Y.~Lecun, and K.~Perlin.
\newblock Real-time continuous pose recovery of human hands using convolutional
  networks.
\newblock {\em ACM Transactions on Graphics}, 33(5):169, 2014.

\bibitem{wang2018dynamic}
Y.~Wang, Y.~Sun, Z.~Liu, S.~E. Sarma, M.~M. Bronstein, and J.~M. Solomon.
\newblock Dynamic graph cnn for learning on point clouds.
\newblock {\em arXiv preprint arXiv:1801.07829}, 2018.

\bibitem{wu20153d}
Z.~Wu, S.~Song, A.~Khosla, F.~Yu, L.~Zhang, X.~Tang, and J.~Xiao.
\newblock 3d shapenets: A deep representation for volumetric shapes.
\newblock In {\em Proceedings of the IEEE conference on computer vision and
  pattern recognition}, pages 1912--1920, 2015.

\bibitem{ye2018occlusion}
Q.~Ye and T.-K. Kim.
\newblock Occlusion-aware hand pose estimation using hierarchical mixture
  density network.
\newblock {\em ECCV, Springer}, 2018.

\bibitem{ye2016spatial}
Q.~Ye, S.~Yuan, and T.-K. Kim.
\newblock Spatial attention deep net with partial pso for hierarchical hybrid
  hand pose estimation.
\newblock In {\em European conference on computer vision}, pages 346--361.
  Springer, 2016.

\bibitem{Yu_2018_CVPR}
T.~Yu, J.~Meng, and J.~Yuan.
\newblock Multi-view harmonized bilinear network for 3d object recognition.
\newblock In {\em The IEEE Conference on Computer Vision and Pattern
  Recognition (CVPR)}, June 2018.

\bibitem{Yuan2017c}
S.~Yuan, G.~Garcia-Hernando, B.~Stenger, G.~Moon, J.~Yong~Chang, K.~Mu~Lee,
  P.~Molchanov, J.~Kautz, S.~Honari, L.~Ge, J.~Yuan, X.~Chen, G.~Wang, F.~Yang,
  K.~Akiyama, Y.~Wu, Q.~Wan, M.~Madadi, S.~Escalera, S.~Li, D.~Lee,
  I.~Oikonomidis, A.~Argyros, and T.-K. Kim.
\newblock Depth-based 3d hand pose estimation: From current achievements to
  future goals.
\newblock In {\em The IEEE Conference on Computer Vision and Pattern
  Recognition (CVPR)}, June 2018.

\bibitem{Yuan2017b}
S.~Yuan, Q.~Ye, G.~Garcia-Hernando, and T.-K. Kim.
\newblock The 2017 hands in the million challenge on 3d hand pose estimation.
\newblock {\em arXiv:1707.02237}, 2017.

\bibitem{Yuan2017a}
S.~Yuan, Q.~Ye, B.~Stenger, S.~Jain, and T.-K. Kim.
\newblock Bighand2. 2m benchmark: Hand pose dataset and state of the art
  analysis.
\newblock In {\em IEEE Conference on Computer Vision and Pattern Recognition},
  pages 2605--2613, 2017.

\bibitem{zaheer2017deep}
M.~Zaheer, S.~Kottur, S.~Ravanbakhsh, B.~Poczos, R.~R. Salakhutdinov, and A.~J.
  Smola.
\newblock Deep sets.
\newblock In {\em Advances in Neural Information Processing Systems}, pages
  3391--3401, 2017.

\bibitem{Zhou2016}
X.~Zhou, Q.~Wan, W.~Zhang, X.~Xue, and Y.~Wei.
\newblock Model-based deep hand pose estimation.
\newblock In {\em Twenty-Fifth International Joint Conference on Artificial
  Intelligence}, pages 2421--2427. AAAI Press, 2016.

\end{thebibliography}
}

\end{document}